\documentclass[letterpaper, 10pt, conference]{IEEEtran}

\IEEEoverridecommandlockouts    

\usepackage[T1]{fontenc}
\usepackage{hyperref}

\usepackage{graphicx} 
\usepackage{amsmath} 
\usepackage{amsfonts}       
\usepackage{nicefrac}       
\usepackage{microtype}      
\usepackage{subfig}			
\usepackage{tikz}
\usepackage{pgfplots}
\usepackage{comment}
\usepackage[noadjust]{cite}

\usepackage{adjustbox}
\usepackage{booktabs}	
\usepackage{multirow}
\usepackage{url}
\usepackage{amsthm}
\usepackage{listings}
\usepackage[linesnumbered,ruled,vlined]{algorithm2e}
\usepackage{xcolor}

\definecolor{blu}{rgb}{0, 0,0.7}
\definecolor{verde}{rgb}{0, 0.7,0}
\definecolor{nero}{rgb}{0, 0,0}
\definecolor{rosso}{rgb}{0.7, 0,0}
\definecolor{grigio}{rgb}{0.3, 0.3,0.3}
\definecolor{bianco}{rgb}{1,1,1}

\newcommand{\fillbox}[3]
{\bgroup
  \dimen1=#1\relax
  \dimen2=#2\relax
  \sbox0{\includegraphics[width=#1]{#3}}%
  \ifdim\ht0>\dimen2
    \dimen0=\dimexpr \ht0-\dimen2\relax
    \adjustbox{clip=true,trim=0pt 0.5\dimen0 0pt 0.5\dimen0}{\usebox0}%
  \else
    \sbox0{\includegraphics[height=#2]{#3}}%
    \ifdim\wd0>\dimen1
      \dimen0=\dimexpr \wd0-\dimen1\relax
      \adjustbox{clip=true,trim=0.5\dimen0 0pt 0.5\dimen0 0pt}{\usebox0}%
    \else
      \usebox0
    \fi
  \fi
\egroup}

\DeclareMathOperator*{\argmax}{argmax}

\title{On Using Neural Networks to Learn Safety Speed Reduction in Human-Robot Collaboration: \\
A Comparative Analysis}

\author{Marco Faroni, Alessio Spanò, Andrea M. Zanchettin, Paolo Rocco
\thanks{This study was partially carried out within the MICS (Made in Italy – Circular and Sustainable) Extended Partnership and received funding from Next-Generation EU (Italian PNRR – M4 C2, Invest 1.3 – D.D. 1551.11-10-2022, PE00000004). CUP MICS D43C22003120001.}
\thanks{The authors are with Politecnico di Milano, Piazza Leonardo da Vinci, 32. Milano (Italy) {\tt\footnotesize marco.faroni@polimi.it}
}
}

\begin{document}

\maketitle
\thispagestyle{empty}
\pagestyle{empty}

\begin{abstract}
In Human-Robot Collaboration, safety mechanisms such as Speed and Separation Monitoring and Power and Force Limitation dynamically adjust the robot’s speed based on human proximity. While essential for risk reduction, these mechanisms introduce slowdowns that makes cycle time estimation a hard task and impact job scheduling efficiency. Existing methods for estimating cycle times or designing schedulers often rely on predefined safety models, which may not accurately reflect real-world safety implementations, as these depend on case-specific risk assessments.
In this paper, we propose a deep learning approach to predict the robot’s safety scaling factor directly from process execution data. 
We analyze multiple neural network architectures and demonstrate that a simple feed-forward network effectively estimates the robot’s slowdown. This capability is crucial for improving cycle time predictions and designing more effective scheduling algorithms in collaborative robotic environments.
\end{abstract}

\section{Introduction}\label{sec:intro}

Human-Robot Collaboration (HRC) is transforming modern industrial environments by enabling robots and humans to work together in shared workspaces. Unlike traditional robotic systems, which operate in isolated areas with predefined tasks, collaborative robots (cobots) are designed to dynamically interact with human workers, assisting in complex and flexible manufacturing processes. This collaboration enhances productivity, adaptability, and ergonomics, reducing the physical and cognitive load on human operators. However, ensuring safety in HRC remains a critical challenge, as robots must adjust their behavior to prevent potential accidents while maintaining efficiency.

To achieve safe interactions, collaborative robots rely on safety mechanisms that regulate their motion based on human presence and proximity. These mechanisms often require speed modulation as described by international standards such as ISO 10218 and ISO TS 15066. While these safety measures are essential, they can introduce dead times—periods where the robot slows down or halts unnecessarily—resulting in decreased productivity and stochastic cycle times. The trade-off between safety and efficiency is a central issue in HRC, and finding an optimal balance is crucial for the widespread adoption of collaborative robotic systems.

To mitigate these inefficiencies, researchers have explored safety-aware control strategies that integrate real-time human motion prediction and adaptive robot behavior \cite{Makris2022}. Many of these approaches assume a predefined safety function, such as a known speed scaling model, to govern how the robot adjusts its motion in response to human presence. However, in practical applications, safety implementations are determined by system integrators based on risk assessments, leading to significant deviations from standardized guidelines. Unlike the idealized continuous speed reduction models described in ISO TS 15066, real-world implementations often rely on static safety zones or binary human detection, making it difficult to apply predefined control strategies effectively.

This gap between theoretical safety models and real-world implementations limits the effectiveness of safety-aware control strategies. In this work, we propose a data-driven approach to overcome this limitation by learning the safety function directly from real-world execution. We develop a deep learning-based regression model that estimates the robot's safety scaling factor based on the current state of both the robot and the human. By evaluating different network architectures, we demonstrate that a simple feed-forward neural network can effectively predict the safety-related slowdown of the robot with minimal prior knowledge of the underlying safety implementation. 

\section{Related work}\label{subsec: related-works}

Ensuring safety in Human-Robot Collaboration (HRC) is a fundamental requirement outlined in international standards such as ISO 10218-2 \cite{ISO10218-2} and ISO TS 15066 \cite{ISOTS15066}. These standards provide general safety guidelines but leave the actual risk assessment and mitigation strategies to system integrators, following the principles of ISO 12100. Among the most widely adopted safety mechanisms, Speed and Separation Monitoring (SSM) and Power and Force Limitation (PFL) dynamically regulate the robot’s speed based on its proximity to humans. While these mechanisms are effective in preventing accidents, they often introduce significant slowdowns, reducing the overall efficiency of collaborative tasks.

Several studies have attempted to mitigate these inefficiencies by incorporating safety-awareness into different levels of the robot’s control architecture spanning scheduling \cite{Faroni_Sandrini_RCIM,Zanchettin:scheduling,faccio2023task}, motion planning \cite{HAMP,Faroni_Tonola_TASE,Haddadin:S_star}, and control \cite{pupa:mpc}.

Despite these advances, existing solutions assume a predefined safety model based on ISO TS 15066 guidelines. In real-world applications, however, safety implementations vary significantly due to case-specific risk assessments, leading to deviations from theoretical models. As a result, safety-aware control strategies based on predefined models may not accurately reflect the actual safety logic deployed in a system, limiting their effectiveness and generalization.

Safety-related downtime is also a major source of stochasticity in the estimation of the process cycle time.
Traditionally, cycle time estimation for robots was based on their motion characteristics (e.g., velocity, acceleration, and travel distance), while human task times were estimated using Method-Time Measurement \cite{KOMENDA20211119}. 
As robots and humans are increasingly integrated into joint workflows, recent approaches have combined these methods to account for both entities' tasks in a holistic way \cite{schroter2016introducing, michalos2018method}. 
These methods are valuable for estimating cycle times in repetitive tasks, assuming human-robot interactions are predictable.
Alternatively, Pellegrinelli and Pedrocchi  \cite{pellegrinelli2017estimation} predict average cycle times in shared human-robot spaces.
They model the occurrence of safety stops as a Markov chain and compute the expected cycle time according to the average human workspace occupancy and the consequent statistical interference between human and robot occupied voxels.
However, their model relies on a known safety logic and the associated downtime.

A related area of research focuses on human motion prediction, which is often integrated into safety-aware control frameworks \cite{hierarchical-human-motion-prediction, uncertainty-hamp, scalera2024robust, uncertain-human-predictio-planning}. Human movement can be predicted using physics-based models \cite{hermann2015anticipate}, filtering techniques \cite{ferrari2024predicting}, or deep learning methods that leverage contextual information for long-term trajectory forecasting \cite{finean2023motion, long-term-3, toussaint-long-term-4}. While these methods improve human-awareness in collaborative scenarios, they do not directly estimate how safety mechanisms influence the robot’s motion.

To the best of our knowledge, no prior works focus on learning the robot’s safety logic directly from data, or how these slowdowns affect cycle times in human-robot systems. This paper addresses this gap by learning the robot’s safety scaling from data, allowing the system to predict safety-induced slowdowns without predefined safety models or expert-driven assumptions. By integrating deep learning techniques, we predict the robot’s slowdown based on real-time process execution data, which improves the accuracy of cycle time estimates in environments with dynamic safety constraints.

\section{Problem Statement}\label{sec:problem}

We consider a collaborative cell composed of one robot and one human working in a shared space. We denote the robot position by $x_r\in  \mathbb R^3$ and the human position by $x_h\in \mathbb R^3$. 
If the human is not present, the robot performs trajectories at the nominal speed; otherwise, the robot moves at a fraction of the nominal speed. 
Such fraction is determined by the safety scaling function $s(x_r,x_h)$, where $s: \mathbb R^6 \mapsto [0,1]$. 
Hence, if $s=0$ the robot halts, while $s=1$ means that the robot moves at the nominal speed. 
We consider staircase safety functions, i.e., $s$ has the following shape:
\begin{equation} \label{eq: safety-staircase}
s(x_r,x_h) = \left\{\begin{array}{ll}
        s_1, & \text{if } \gamma(x_r,x_h) \in D_1\\
        s_2, & \text{if } \gamma(x_r,x_h) \in D_2\\
        \dots  & \\
        s_P, & \text{if } \gamma(x_r,x_h) \in D_P
        \end{array}\right.
\end{equation}
where $\{s_1,\dots,s_P\}$ are constant scaling values with $s_i < s_{i+1}$, $\gamma(x_r,x_h)$ is a generic function in the robot and human positions, and $D_i$ is a generic set whose membership determines the scaling value.

This situation resembles the typical implementation of SSM and PFL according to ISO 10218-2, where safety values are applied according to a discretization of the human and robot positions (e.g., safety areas).

Our approach assumes that (i) the current values of $x_r$ and $s$ can be measured with negligible uncertainty and (ii) the current value of $x_h$ can be measured with uncertainty so that $x_h \sim N(\mu_h,\sigma_h)$. 
The assumption on $x_r$ is reasonable for industrial robots, which typically ensure a precision smaller than 0.1 mm and an accuracy smaller than 1 mm. 
The assumption on $x_h$ comes from the fact that the current technologies for human tracking (e.g., radar, laser scanners, cameras) come with uncertainty up to a few centimeters. 
As for $s$, most robots explicitly provide the speed override value. 
In general, it is possible to retrieve the robot speed scaling by computing the ratio between the measured velocity and the velocity obtained from a run without the human presence.

We also assume that the human and robot's movements can be associated with a human and robot's goals, $g_h$ and $g_r$, respectively. 
For example, $g_r$ is the final position of the robot's ongoing trajectory, while $g_h$ is the operator's expected position at the end of the ongoing task.

As we do not have access to the true values of $s_i$, $D_i$, and $\gamma$, we aim to estimate the safety speed function \eqref{eq: safety-staircase} from data collected during the process execution. 
In particular, we consider the following problems:
\begin{enumerate}
\item given the current $x_r$ and $x_h$, predict the current scaling value;
\item given the current $x_r$, $x_h$, $g_r$, and $g_h$, predict the scaling value at a future time instant;
\item given the current $x_r$, $x_h$, $g_r$, and $g_h$, predict the average scaling value over a future time horizon;
\end{enumerate}

\section{Methods}\label{sec: method}

This section compares different neural network architectures to solve the problems above. 
First, we discuss how to estimate the number of steps, $P$, in $s$, which will be beneficial to the proposed networks. 
Then, we describe the networks proposed for each prediction problem.

\subsection{Data Collection and Self-Labeling via Clustering}

The neural networks are trained with a collection of process executions. 
We consider a dataset $\mathcal{D}=(D_1, \dots, D_{M})$, where $D_i$ is a tuple containing the robot and human states and goals, and the scaling value at the $i$th time step, 
\begin{equation}
\label{eq: data-point}
   D_i=\big(\hat{x}_{r, i}, \hat{x}_{h, i}, \hat{g}_{r, i}, \hat{g}_{h, i}, \hat{s}_i \big)
\end{equation}
where $\hat{x}_{r, i}$, $\hat{x}_{h, i}$, $\hat{g}_{r, i}$, $\hat{g}_{h, i}$, and $\hat{s}_i$ are the $i$th observations of $x_r$, $x_h$, $g_r$, $g_h$, and $s$. 

Note that we do not assume the number $P$ of steps in $s$ is known. 
We can approximate $P$ by the number of clusters obtained by applying a clustering algorithm on the collected values of $\{\hat{s}_i\}_{i=1}^M$. 
Most clustering algorithms automatically output the optimal number of clusters. 
For those that do not (e.g., K-means), $P$ can be computed using the Silhouette or the elbow method \cite{Silhouettes}. 

The clustering algorithm returns a set of clusters $\mathcal{C}=\{C_1,...,C_P\}$ and assigns each point $\hat{s}$ to a cluster.
We define an augmented tuple as:
\begin{equation}
\label{eq: data-point-augmented-1}
D_i=\big( \hat{x}_{r, i}, \hat{x}_{h, i}, \hat{g}_{r, i}, \hat{g}_{h, i}, \hat{s}_i, C(\hat{s}_i ) \big)
\end{equation}
where $C\big( \hat{s}_i \big)$ is the cluster associated with $\hat{s}_i$.
Finally, each cluster is associated with a speed scaling value equal to the centroid of the cluster, $c_j$, so that
\begin{equation}
\label{eq: cluster-centroid}
    c_j = \frac{\sum_{i=1}^M \mathbf{I}\big(C(\hat{s}_i=C_j)\big) \hat{s}_i}{\sum_{i=1}^M \mathbf{I}\big(C(\hat{s}_i)=C_j\big)} 
\end{equation}
where $\mathbf{I}(\cdot)$ is the indicator function.

\begin{figure}[tpb] \label{fig: networks}
    \centering
    \setlength{\unitlength}{0.1\columnwidth}
    \subfloat[Classification network (Sec. \ref{sec: one-step})]{
        \begin{picture}(10,4.5)
            \put(1.2,0.2){\includegraphics[trim=0cm 9cm 0cm 0cm, clip, width=0.77\columnwidth]{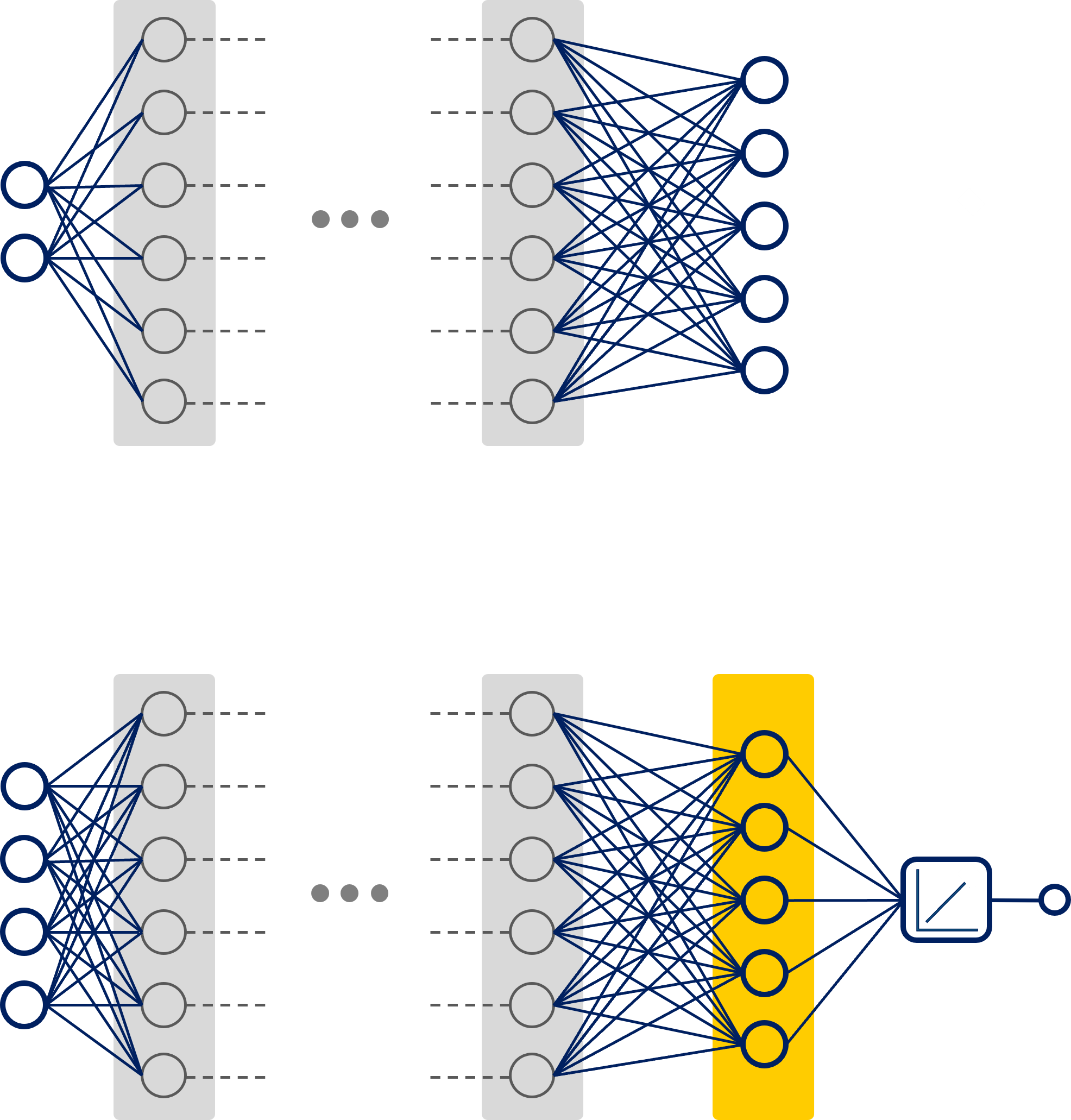}}
            \put(0.7,2.3){$x_r$}
            \put(0.7,1.8){$x_h$}
            \put(7.1,2){$\tilde{y}\in \mathbb R^P$}
            \put(1.9,4){$\overbrace{\qquad\qquad\qquad\qquad\quad}^\text{hidden layers}$}
        \end{picture}
        \label{fig: networks-1}
    }\\
    \subfloat[Mixed approach (Sec. \ref{sec: avg-prediction})]{
        \begin{picture}(10,4.5)
            \put(1.2,0.2){\includegraphics[trim=0cm 0cm 0cm 9cm, clip, width=0.77\columnwidth]{img/nn1.png}}
            \put(0.7,2.6){$x_r$}
            \put(0.7,2.0){$x_h$}
            \put(0.7,1.5){$g_r$}
            \put(0.7,0.9){$g_h$}
            \put(9,1.65){$\tilde{y}$}
            \put(1.8,4){$\overbrace{\qquad\qquad\qquad\qquad\qquad\qquad\qquad}^\text{hidden layers}$}
            \put(6.2,3.6){\footnotesize width=$P$}        
            \put(7.7,2.2){\footnotesize linear}        
        \end{picture}
        \label{fig: networks-2}
    }
    \caption{Neural network architecture.}
    \label{fig: network}
\end{figure}

\subsection{One-step scaling prediction}\label{sec: one-step}

As $s$ is a staircase function, we frame the learning problem as a regression or a multi-class classification problem:

\subsubsection{Classification network}

We can see each step in $s$ as a class of our classification problem. 
We can consider the cluster membership of each data point \eqref{eq: data-point-augmented-1} as a class label.
Specifically, we associate each cluster, $C_i,$ to a one-hot encoded label vector, $l(C_i) = (l_{1},...,l_j,...,l_{P})$, such that:
\begin{equation}
    l_{j} = \left\{
    \begin{array}{ll}
        1, & \text{if } i=j\\
        0, & \text{otherwise}\\
    \end{array}\right.
\end{equation}

We use a feed-forward deep neural network to solve the resulting $P$-class classification problem where:
\begin{itemize}
    \item The input vector is $x = (x_r,x_h) \in \mathbb R^6$;
    \item The output $\tilde{y} \in \mathbb R^P$ is the probability of each class label (i.e., the output layer uses a \texttt{Softmax} activation function);
    \item The loss function $L$ is the Cross Entropy function
    $$
        L\Big( x,l\big( C(\hat{s}) \big) \Big) = -\sum_{j=1}^P l_j\big( C(\hat{s}) \big) \log\big( \tilde{y}_j({x}) \big)
    $$
\end{itemize}

Finally, the predicted scaling value, $\tilde{s}$, is the centroid of the predicted cluster label, i.e.,
\begin{equation}
    \tilde{s} = c_k \text{ with } k = \argmax_i \tilde{y}_i
\end{equation}
An example of the resulting network for $P=5$ is in Fig. \ref{fig: networks-2}.

\subsubsection{Regression} 
We aim to directly learn the function $s$ by using a regressor. 
We use a feed-forward neural network with input ${x}=(x_r,x_h)$, output $\tilde{y} \in \mathbb R$, \texttt{Hardtanh} output activation function, and Mean Squared Error (\texttt{MSE}) loss function:
\begin{equation}
    L(x,\hat{s}) = || \hat{s} - \tilde{y}(x) ||^2
\end{equation}



\subsection{N-step scaling prediction}\label{sec: N-step}

We aim to predict $s$ at time $t+w\Delta T$, where $w\in \mathbb N$ and $\Delta T$ is the data collection sampling period. 
The problem is similar to the previous one except the input is $x =(x_r, x_h, g_r, g_h)$. 
Conditioning the prediction on the human and robot's goals significantly improves prediction accuracy, as the goals inform the network of the agents' intentions for future movements. 
The three networks proposed in Sec. \ref{sec: one-step} can be used without changes except that the input layer has a width equal to 12 and the training data point becomes:
\begin{equation}
    D_i'=\big( \hat{x}_{r, i}, \hat{x}_{h, i}, \hat{g}_{r, i}, \hat{g}_{h, i}, \hat{s}_{i+w}, C( \hat{s}_{i+w} ) \big)
\end{equation}
where $\hat{s}_{i+w}$ denotes the observed scaling value at time $t+w\Delta T$. 


\subsection{Average scaling prediction}\label{sec: avg-prediction}

We aim to predict the average scaling over a time window $[t,t+\Delta T w]$. 
The classification problem does not apply in this case, as the output function can take any value between 0 and 1. 
Nonetheless, we can build on the classification network to heuristically inform the design of a regression network.

To this purpose, note that the average scaling is a linear combination of the values $\{s_1,\dots,s_P\}$ so that
\begin{equation}
    \tilde{y} = \sum_i^P \alpha_i s_i \text{ with } \sum_i^P \alpha_i=1
\end{equation}
where $\alpha_i$ represents the fraction of the interval $[t,t+w\Delta T]$ where $\hat{s}=s_i$. 
The output of the classification network is a vector, $\tilde{y} \in [0,1]^P$ such that $\sum_{i=1}^P \tilde{y}_i = 1$, because of the \texttt{Softmax} output function. 

We draw upon this analogy to re-use the structure of the classification network in a regression network by adding a linear output layer.
The output of the resulting network is
\begin{equation}
    \tilde{y}'=\beta_0 + \sum_{i=1}^P \beta_i \tilde{y}_i
\end{equation}
where $\beta \in \mathbb R^{P+1}$ is the weight vector of the last layer. 
We can therefore expect the first part of the network to learn a feature vector proportional to the fraction of the interval $[t,t+w\Delta T]$ where $\hat{s}=s_i\forall i$, while the last linear layer will weigh each contribution and yield the average scaling value.

We use the \texttt{MSE} loss function to train the network. 
To do so, the training data points become
\begin{equation}
    D_i''=\big( \hat{x}_{r, i}, \hat{x}_{h, i}, \hat{g}_{r, i}, \hat{g}_{h, i}, \bar{s}_{i+w} \big)\end{equation}
where
\begin{equation}
    \bar{s}_i = \frac{1}{w+1} \sum_{j=0}^{w}  \hat{s}_{i+j} ,
\end{equation}
the input vector is $x=(\hat{x}_{r}, \hat{x}_{h}, \hat{g}_{r}, \hat{g}_{h})$, and the loss function is $L=|| \bar{s} - \tilde{y}(x) ||^2$.
The resulting network architecture for $P=5$ is exemplified in Fig. \ref{fig: networks-2}.

\section{Experiments}\label{sec: experiments}

\begin{figure*}[tpb]
    \centering
    \includegraphics[trim={0cm 0cm 2cm 0}, clip, height=0.12\textheight]{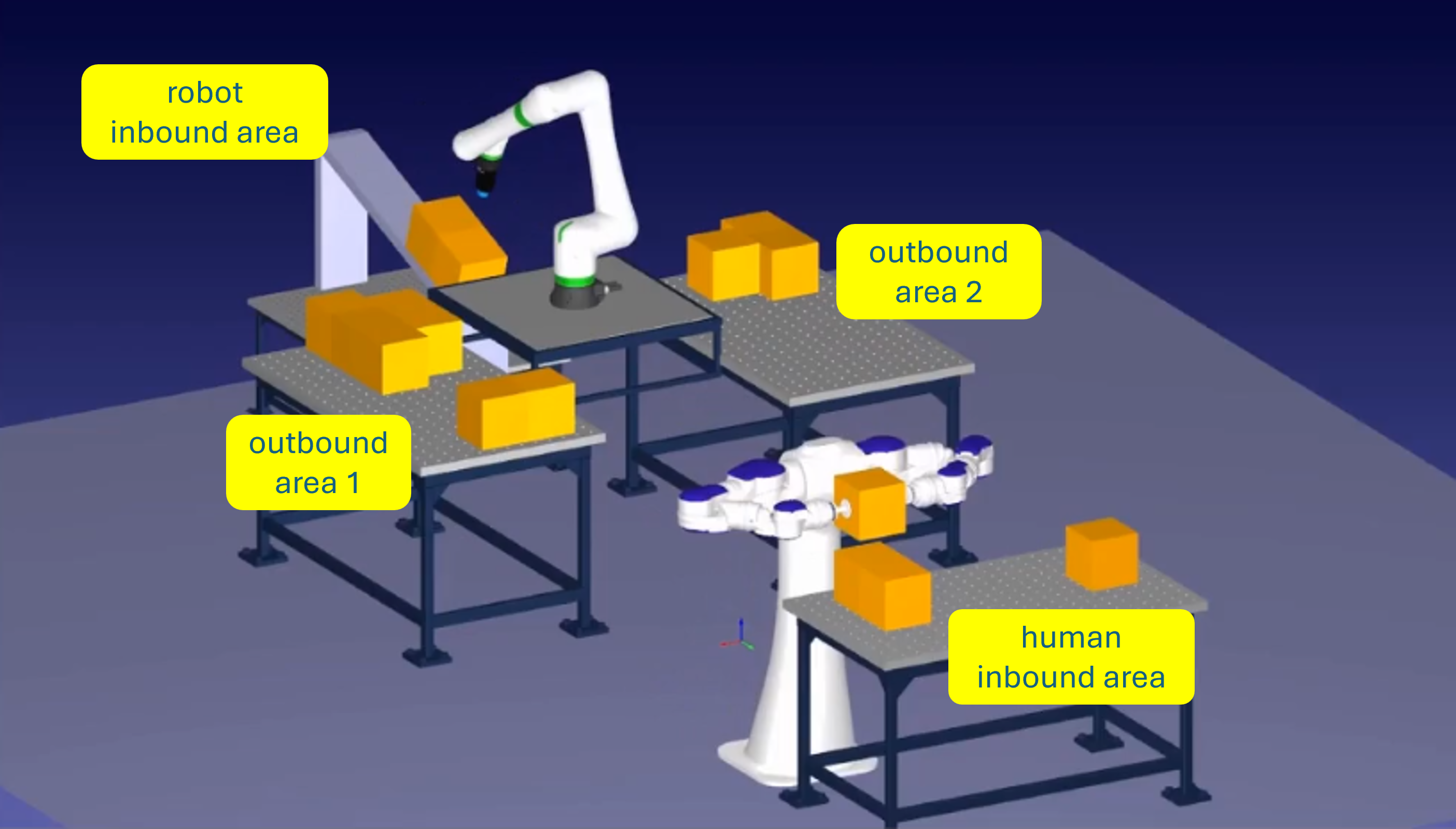} \,
    \includegraphics[height=0.12\textheight]{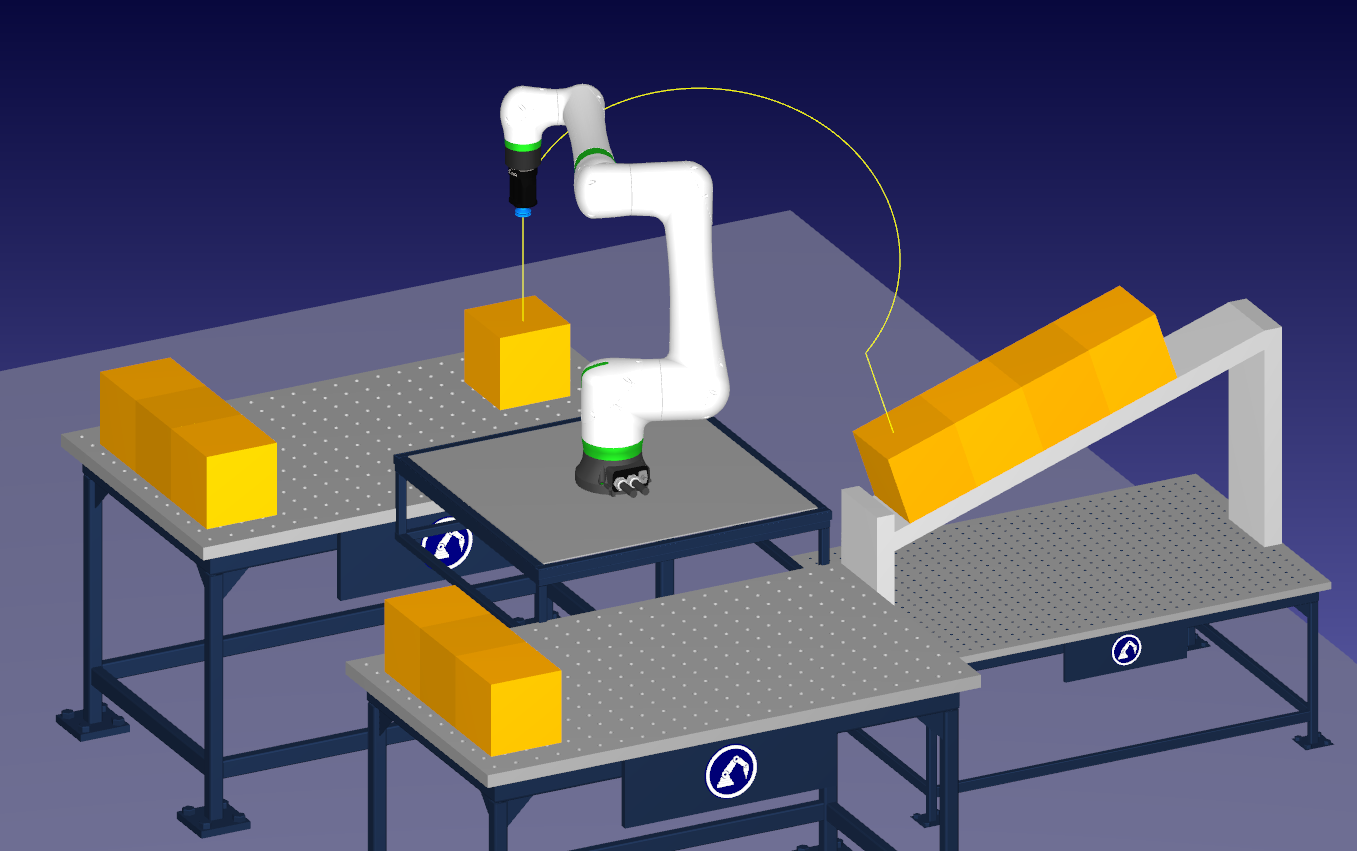} \,
    \includegraphics[height=0.12\textheight]{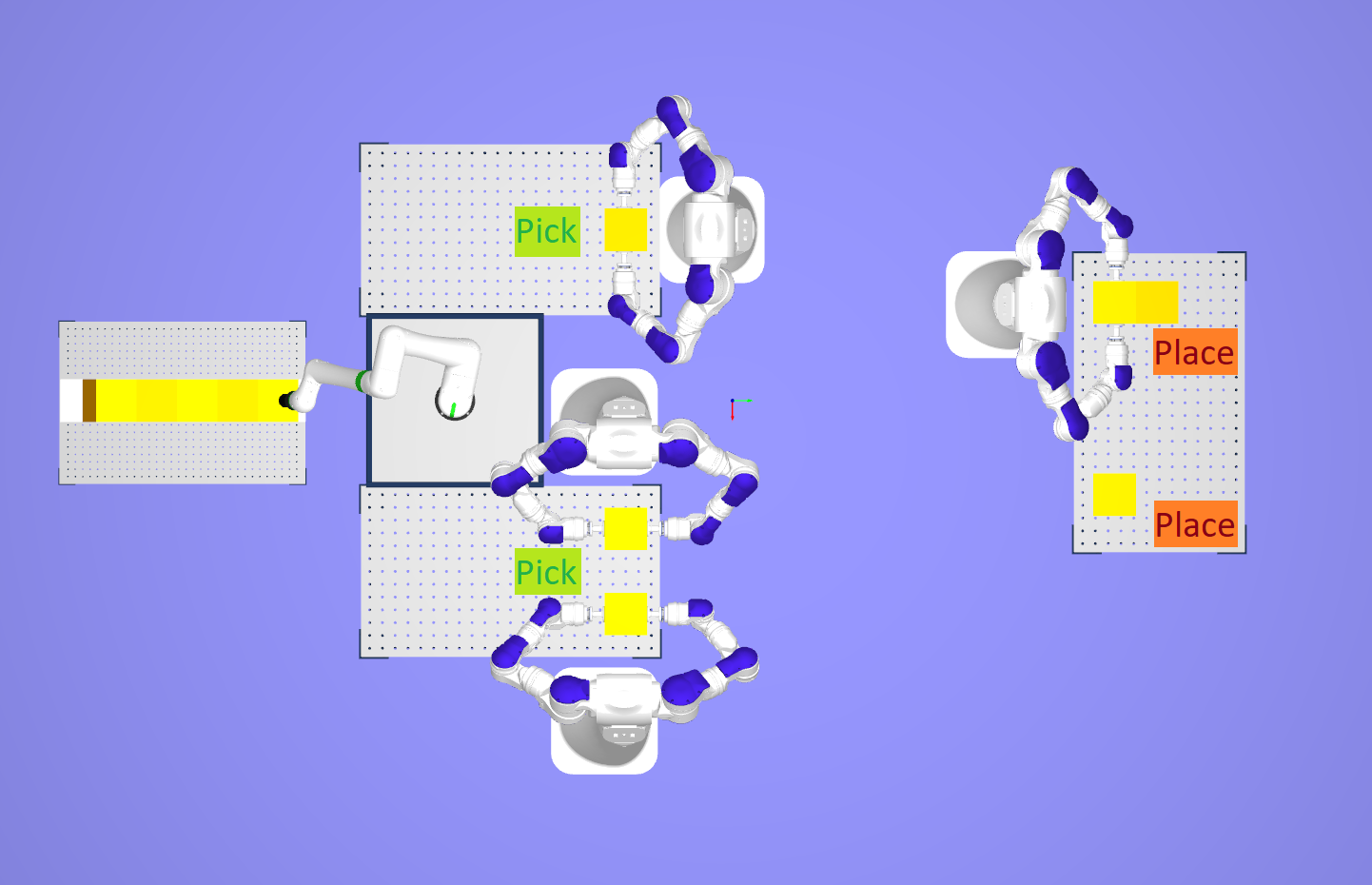}
    \caption{Simulation scenario. Overview of the working space (left). Robot pick\&place example (middle). Human picking and placing nominal positions (right).}
    \label{fig: simulation-scenario}
    \vspace{-0.3cm}
\end{figure*}

We simulate a box-picking scenario using RoboDK, as illustrated in Fig. \ref{fig: simulation-scenario}.
Since RoboDK does not include a built-in human simulator, we represent the operator using a Motoman SDA10F humanoid robot mounted on an omnidirectional mobile base.
The robotic manipulator used in the scenario is a 6-DOF Fanuc CRX10iA equipped with a vacuum gripper.

\begin{figure}
    \centering
    \begin{tikzpicture}
        \begin{axis}[
            width=0.9\columnwidth,
            height=0.45\columnwidth,
            axis lines=middle,
            xlabel style={at={(axis description cs:0.5,-0.25)}, anchor=north},
            xlabel={human-robot distance},
            ylabel={$s$},
            xtick={1,2,3,4},
            xticklabels={$d_1$, $d_2$, $d_3$, $d_4$},
            ytick={0.25, 0.5, 0.75, 1},
            ymin=0, ymax=1.2,
            xmin=0, xmax=5,
            domain=0:4,
            samples=200,
            grid=both,
            grid style={dashed, gray!30}
        ]
        \addplot[
            red,
            thick,
            domain=0:1
        ] {0};
        \addplot[
            red,
            thick,
            domain=1:2
        ] {0.25};
        \addplot[
            red,
            thick,
            domain=2:3
        ] {0.5};
        \addplot[
            red,
            thick,
            domain=3:4
        ] {0.75};
        \addplot[
            red,
            thick,
            domain=4:5
        ] {1};
        \end{axis}
    \end{tikzpicture}
    \caption{Safety speed scaling function. Simulations: $d_1$=1.2 m, $d_2$=1.5 m, $d_3$=1.9 m, $d_4$=2.1 m. Real-world: $d_1$=0.6 m, $d_2$=0.8 m, $d_3$=1.2 m, $d_4$=1.6 m.}
    \label{fig: safety-function}
\end{figure}
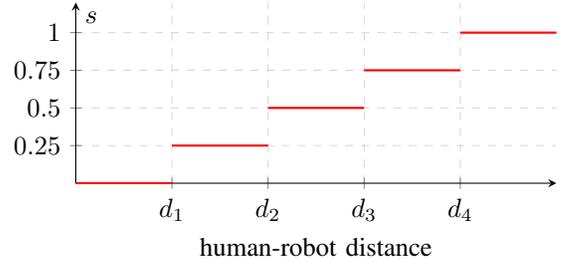

In this setup, both the robot and the operator are responsible for transferring boxes between inbound and outbound areas.
The robot's inbound area is a conveyor belt, while its outbound areas consist of two tables positioned on either side, each capable of holding up to five boxes.
On the other hand, the operator’s inbound area contains six boxes, and their outbound areas—located near the robot—can accommodate up to three boxes each.
The ground-truth safety function $s$ used in this experiment is depicted in Fig. \ref{fig: safety-function} and can be written according to \eqref{eq: safety-staircase} with $\gamma=||x_r-x_h||$ and:
\begin{equation}
\begin{gathered}
    \{s_1, s_2, s_3, s_4, s_5 \} = \{ 0, 0.25, 0.5, 0.75, 1 \} \\
    D_1 = [0,1.2], \,\, 
    D_2 = (1.2,1.5], \,\, 
    D_3 = (1.5,1.9], \\
    D_4 = (1.9,2.4], \,\, 
    D_5 = (2.4,+\infty)
\end{gathered}    
\end{equation}

\subsection{Data collection}

We execute the process 1000 times, recording data at a frequency of 10 Hz for each time step \(i\). The collected data includes:  
\begin{itemize}  
    \item the robot’s end-effector position, \( \hat{x}_{r, i} \in \mathbb{R}^3 \);  
    \item the human’s centroid position, \( \hat{x}_{h, i} \in \mathbb{R}^3 \);  
    \item the robot’s goal position, \( \hat{g}_{r, i} \in \mathbb{R}^3 \);  
    \item the human’s goal position, \( \hat{g}_{h, i} \in \mathbb{R}^3 \);  
    \item the robot’s speed scaling factor, \( \hat{s}_{i} \in [0,1] \).  
\end{itemize}  
To capture the stochasticity of human movements, we randomize the operator's path. Specifically, the human’s goal position is sampled from a Gaussian distribution centered at the nominal goal, with a standard deviation of 0.05 m in the horizontal plane. Additionally, the midpoint of the path is drawn from a Gaussian distribution centered at a nominal midpoint, with a standard deviation of 0.25 m in the horizontal plane.
We use an 80\%-20\% training/test split of the  dataset. 

\subsection{Results} \label{sec: results}

\begin{table}[]
    \caption{\texttt{MSE} of the classification and regression networks for different values of the human measurement noise $\delta$.}
    \centering
    \begin{tabular}{lcccc}
        \toprule
                        & \multicolumn{4}{c}{\texttt{MSE} $\cdot 10^4$ }\\
                        & $\delta = 0\ m $ & $\delta = 0.02\ m $ & $\delta = 0.05\ m $ &  avg. \\
        \midrule
        \footnotesize Classification Network & 3.32 & 8.23 & 21.7 & 11.1 \\
        \footnotesize Regression network     & 2.59 & 7.01 & 15.2 & 8.27 \\
        \bottomrule
    \end{tabular}
    \label{tab: exp-results}
\end{table}

We discuss the results of the proposed networks for the three learning problems described in Sec. \ref{sec: method}.

\subsubsection{One-step scaling prediction}\label{sec: results-one-step}

We instantiate the networks described in Sec. \ref{sec: one-step} as follows:
\begin{itemize}
    \item Classification network: 
    four 64-neuron wide hidden layers with 1D normalization and \texttt{ReLU} activation function. We use the \texttt{DBSCAN} algorithm \cite{dbscan} for the self-labeling clustering.
    \item Regression network:
    five 64-neuron wide hidden layers with  1D normalization and \texttt{ReLU} activation function.
\end{itemize}
The depth and the width of the networks were obtained experimentally as a balance between the accuracy and complexity of the model.

To simulate the effect of measurement noise on the prediction, we compare the networks' \texttt{MSE} value with different noise levels on the human position measures. 
To do so, we add a zero-mean Gaussian noise to the observed values of $x_h$ and evaluate the networks' performance for different variance values. The results are in Table \ref{tab: exp-results}.

\begin{figure}[tpb]
    \includegraphics[trim={0cm 0cm 0cm 25cm}, clip, width=0.99\columnwidth]{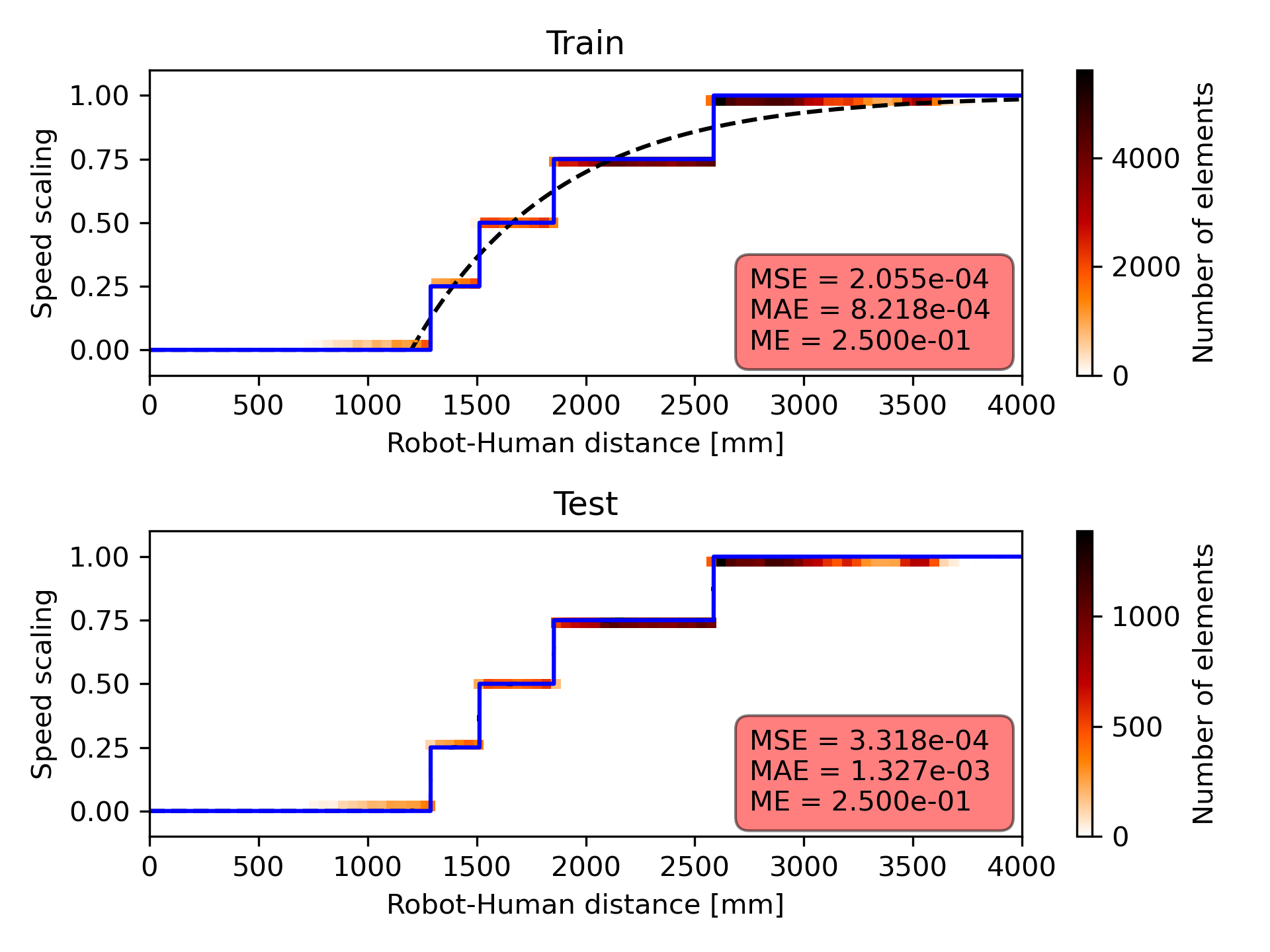}
    \caption{Accuracy of the classification network.}
    \label{fig: classification-accuracy}
\end{figure}

Both networks learn the safety function with satisfactory \texttt{MSE} values but the classification network consistently outperforms the regression one despite its simpler structure and the fact that it was trained with a classification loss function. 
Figure \ref{fig: classification-accuracy} shows the predicted values of the classification network superimposed to the true safety function, showing good prediction capability, even in the neighborhood of the function steps. 

\subsubsection{N-step scaling prediction}\label{sec: results-N-step}

We consider the classification and regression networks of Sec. \ref{sec: N-step}. The networks have the same inner structure as those described in Sec. \ref{sec: results-one-step}. 

\begin{figure}[tpb] 
    \centering
    \setlength{\unitlength}{0.1\columnwidth}
    \subfloat[LSTM-1 network]{
        \begin{picture}(10,3)
            \put(1.2,0.2){\includegraphics[trim=0cm 0cm 12cm 0cm, clip, width=0.65\columnwidth]{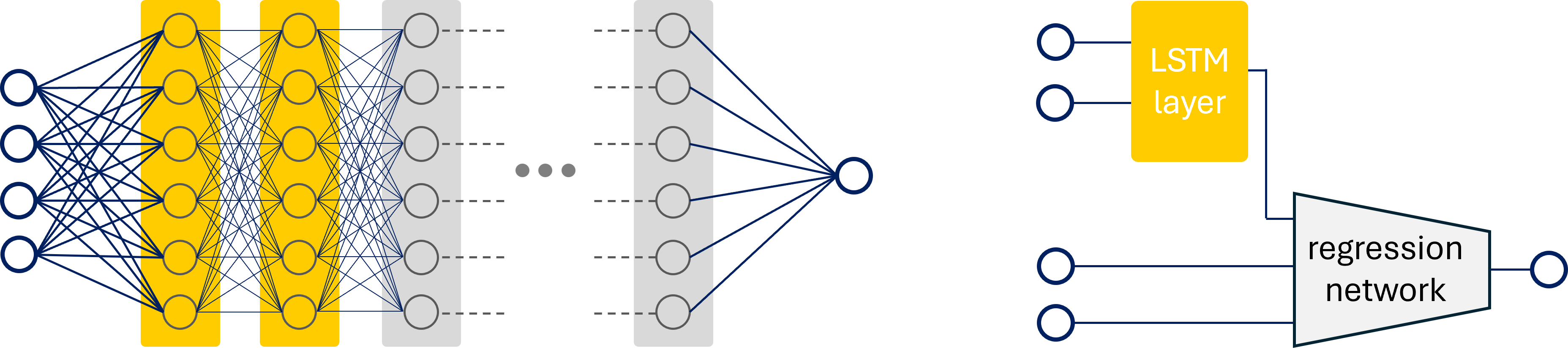}}
            \put(0.7,2){$x_r$}
            \put(0.7,1.6){$x_h$}
            \put(0.7,1.2){$g_r$}
            \put(0.7,0.8){$g_h$}
            \put(7.5,1.4){$\tilde{y}\in \mathbb R$}
            \put(2.1,2.8){$\overbrace{\qquad\qquad}^\text{LSTM layers}$}
        \end{picture}
        \label{fig: networks-lstm1}
    }\\
    \subfloat[LSTM-2 network]{
        \begin{picture}(10,3)
            \put(3.2,0.2){\includegraphics[trim=18cm 0cm 0cm 0cm, clip, width=0.4\columnwidth]{img/nn_lstm.png}}
            \put(2.9,2.3){$x_r$}
            \put(2.9,1.9){$x_h$}
            \put(2.9,0.7){$g_r$}
            \put(2.9,0.3){$g_h$}
            \put(7.5,0.65){$\tilde{y} \in \mathbb R$}        
        \end{picture}
        \label{fig: networks-lstm2}
    }
    \caption{LSTM-based networks for $N$-step predictions.}
    \label{fig: networks-lstm}
\end{figure}

We also consider the two LSTM networks shown in Fig. \ref{fig: networks-lstm} for comparison. 
The first LSTM network substitutes the first two layers of the regression network with LSTM layers. 
The second LSTM network uses a one-layer LSTM network to predict the future human position and plugs it into the regression network. 
The choice of LSTM networks is motivated by their effectiveness at learning temporal patterns, which makes them a preferred choice for prediction over a time horizon.

\begin{figure}[tpb]
    \includegraphics[trim={11.5cm 0cm 0cm 0cm}, clip, width=0.48\columnwidth]{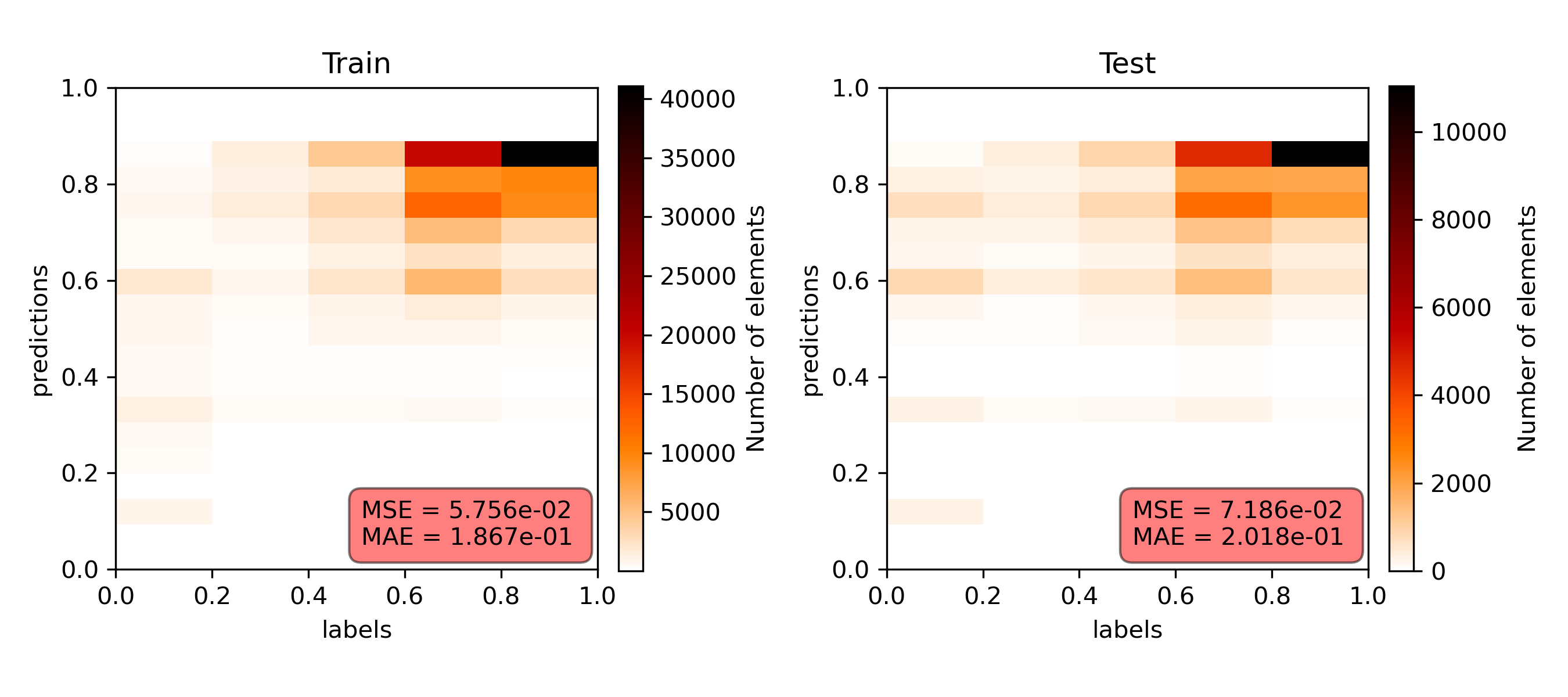}
    \includegraphics[trim={11.5cm 0cm 0cm 0cm}, clip, width=0.48\columnwidth]{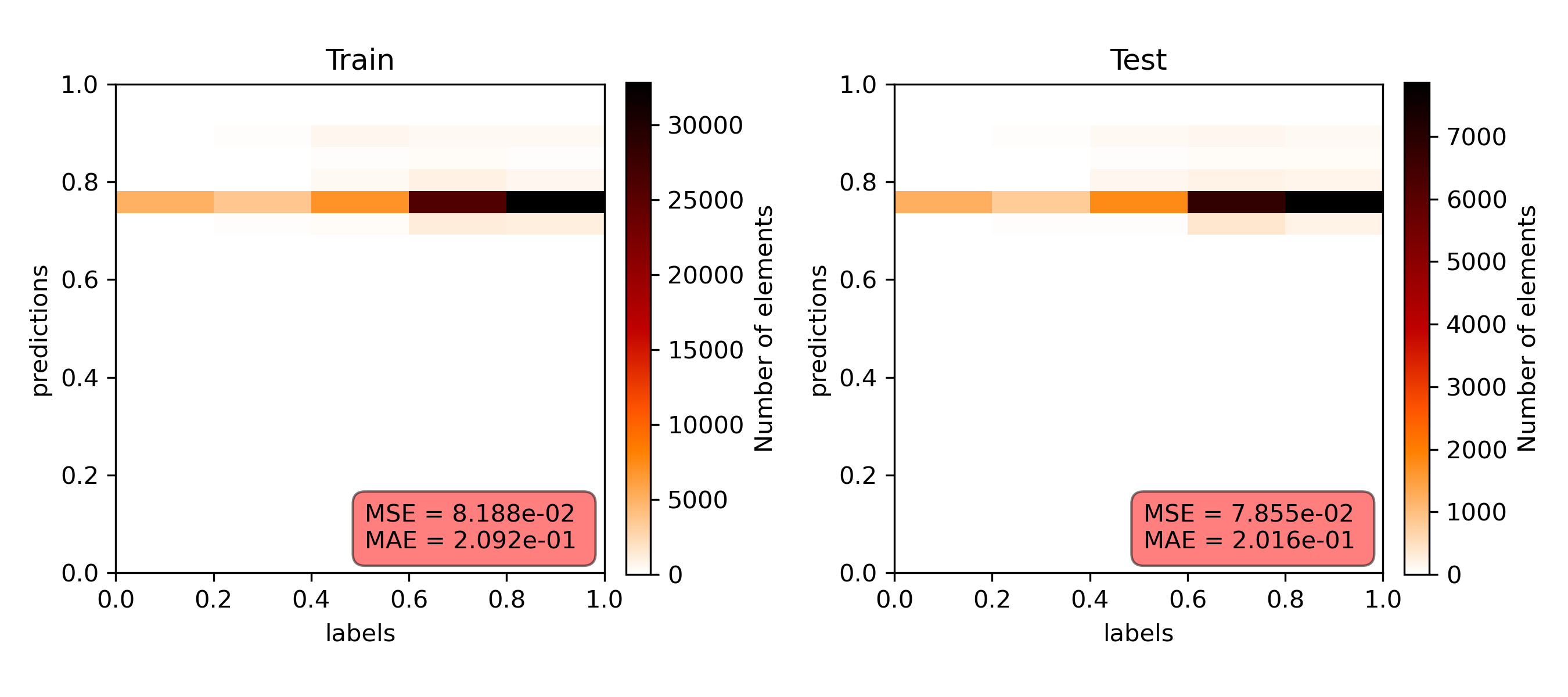}
    \caption{Accuracy heatmap (on test set) of the LSTM-based networks.}
    \label{fig: sim-results-heatmap-lstm}
\end{figure}

The results are in Table \ref{tab: exp-results-lstm}. 
The classification and regression networks perform consistently with the one-step case. 
The LSTM network yields poor prediction results. 
Fig. \ref{fig: sim-results-heatmap-lstm} shows the accuracy of the  LSTM networks with $w\Delta T = 2$ seconds, showing that the network fails at predicting the correct scaling value. 
Such poor results are probably due to the small size of the dataset compared to the needs of an LSTM network. 

\subsubsection{Average scaling prediction}\label{sec: results-avg-prediction}

We consider the network proposed in Sec. \ref{sec: avg-prediction} with 6 hidden layers, where the first 4 layers have a width equal to 64, and the last hidden layer has a width equal to 5, i.e., the number of steps in $s$, retrieved through the clustering algorithm. 

\begin{figure}[tpb]
    \includegraphics[trim={11.5cm 0cm 0cm 0cm}, clip, width=0.48\columnwidth]{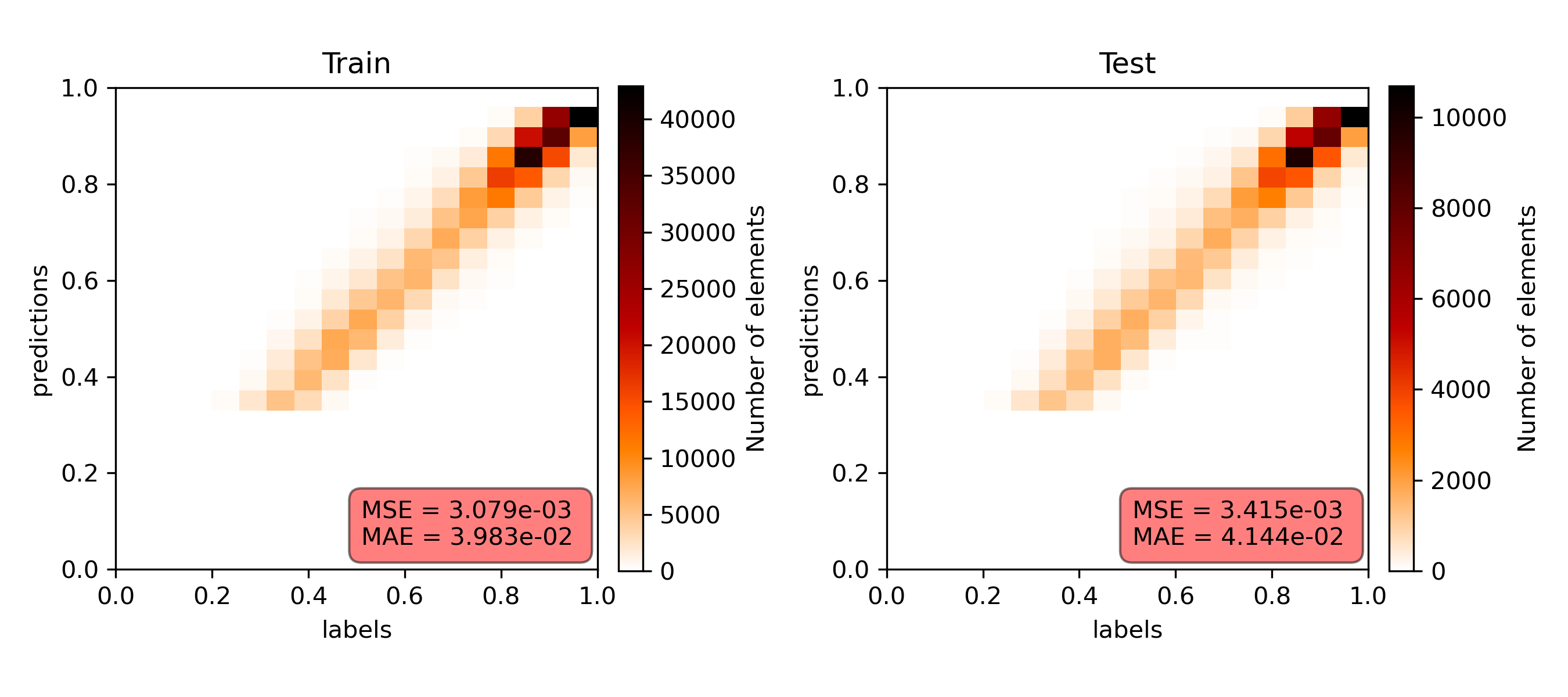}
    \includegraphics[trim={11.5cm 0cm 0cm 0cm}, clip, width=0.48\columnwidth]{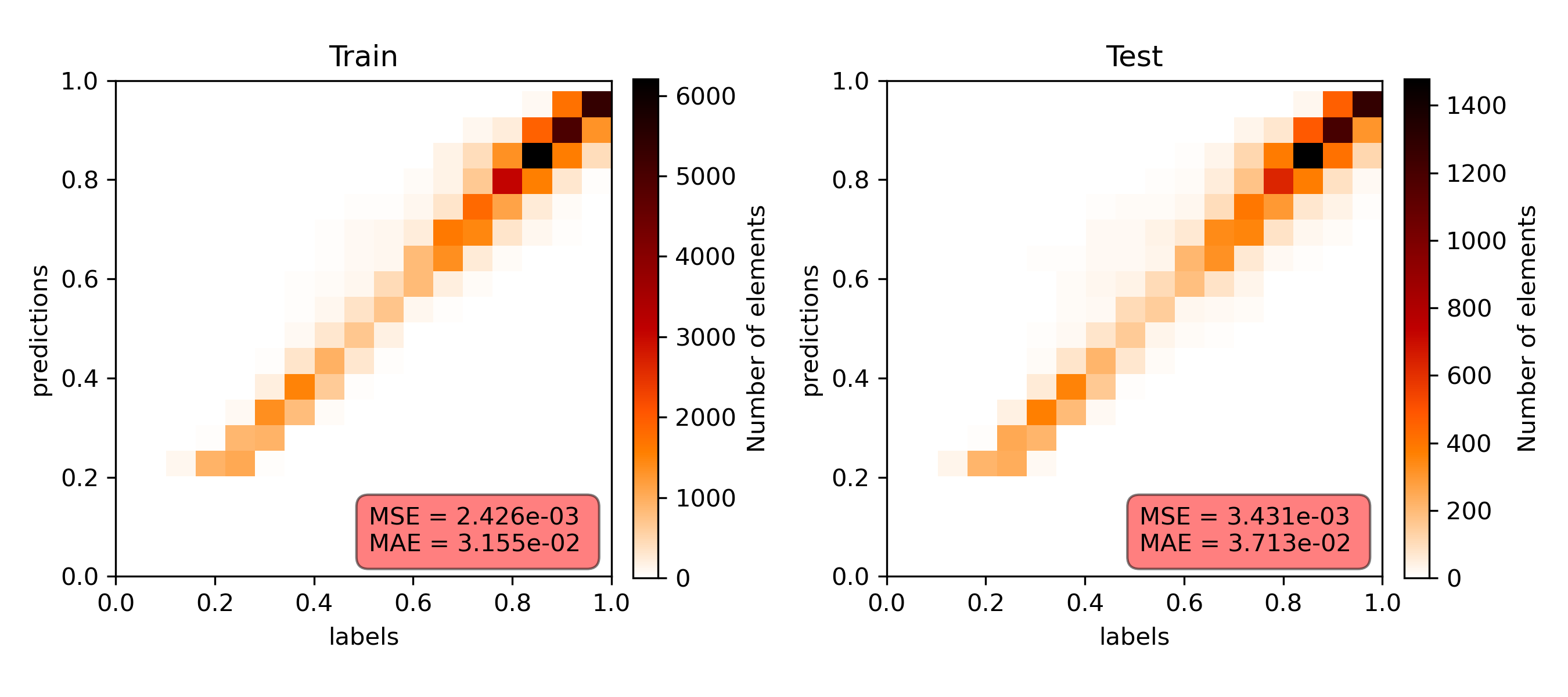}
    \caption{Accuracy heatmap (on test set) of the mixed-approach network for $w\Delta T=14$ s (left) and $19$ s (right).}
    \label{fig: sim-results-heatmap}
\end{figure}

Fig. \ref{fig: sim-results-heatmap} shows the \texttt{MSE} accuracy for $w\Delta T=14$ and 19 seconds. 
The results confirm that the network can effectively learn the average scaling factor over long time intervals.

\section{Real-world Use Case}\label{sec: use-case}

We demonstrate the learning approach in a pick\&packaging scenario. 
The robot picks boxes from a conveyor and places them into the outbound boxes on the table. 
The choice of the outbound box is random. 
At the same time, the operator performs operations such as inspections, packing, and re-filling. 
The robot slows down according to safety function \eqref{eq: safety-staircase} with 
\begin{equation}
\begin{gathered}
    \{s_1, s_2, s_3, s_4, s_5 \} = \{ 0, 0.25, 0.5, 0.75, 1 \} \\
    D_1 = [0,0.6], \,\, 
    D_2 = (0.6,0.8], \,\, 
    D_3 = (0.8,1.2], \\
    D_4 = (1.2,1.6], \,\, 
    D_5 = (1.6,+\infty).
\end{gathered}    
\end{equation}

We collect data from 90 minutes of the process execution with a frequency of 10 Hz. 
The robot position, $x_r$, is the robot tool center point and the robot goal, $g_r$, is the final position of each trajectory. 
The human position, $x_h$, is the centroid of the operator measured by a skeleton tracking system fed by RGB-D images from an Intel Realsense D435i camera. 
The operator communicates its goal, $g_h$, by pressing a button. 

For the sake of brevity, we only consider the problem of learning the average scaling (see Sec. \ref{sec: avg-prediction}). 
The neural network structure is the same as in Sec. \ref{sec: results-avg-prediction}. 

\begin{figure}[tpb]
    \includegraphics[trim={11.5cm 0cm 0cm 0cm}, clip, width=0.48\columnwidth]{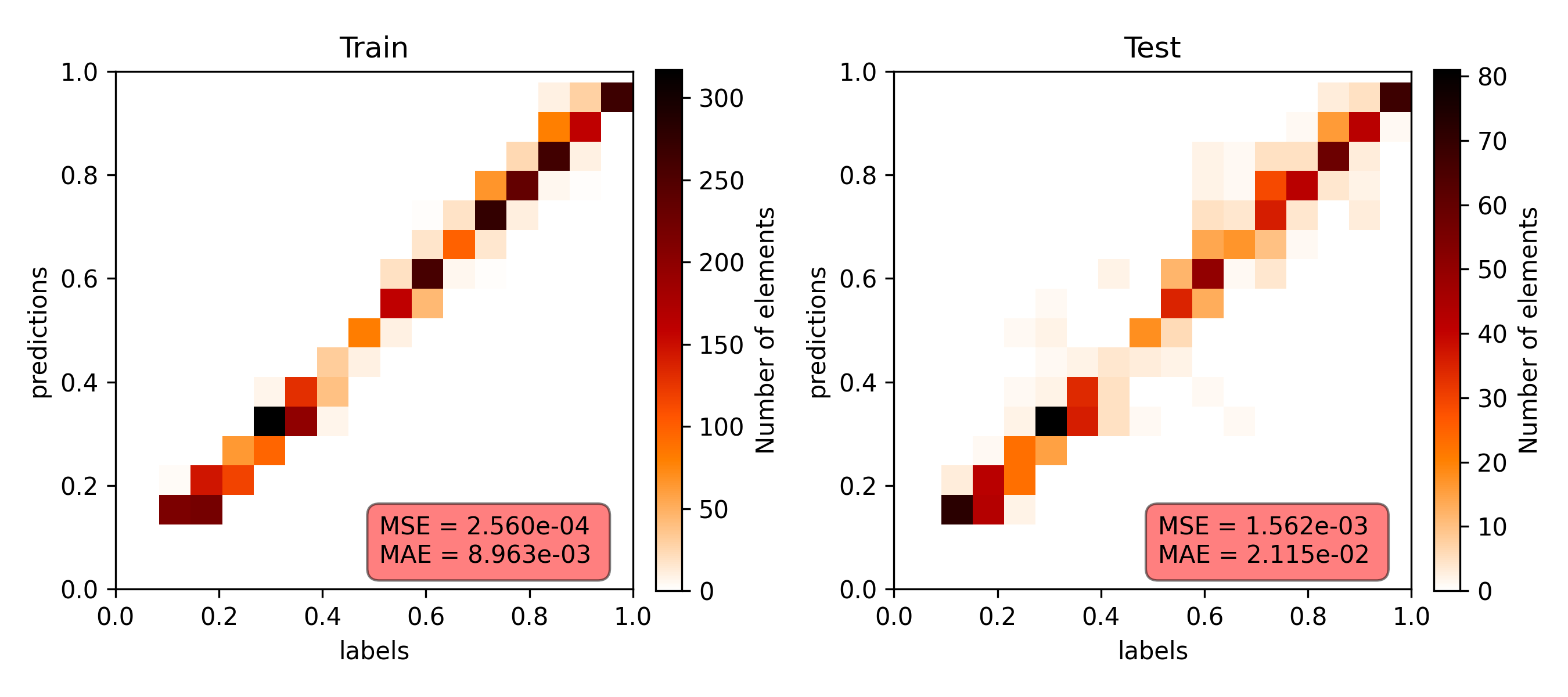}
    \includegraphics[trim={11.5cm 0cm 0cm 0cm}, clip, width=0.48\columnwidth]{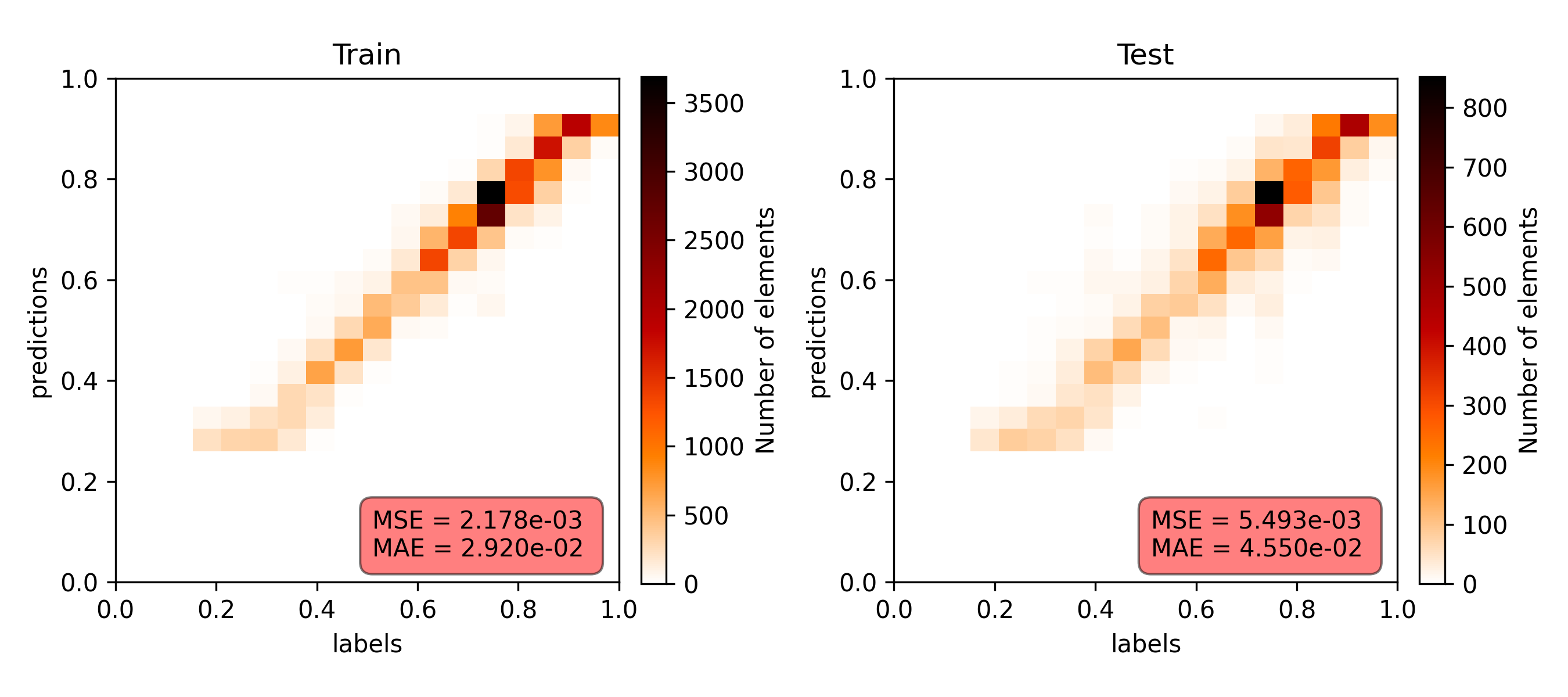}
    \caption{Accuracy heatmap (on test set) of real-world experiments for $w\Delta T=14$ s (left) and $19$ s (right).}
    \label{fig: exp-results-heatmap}
\end{figure}

The results in Fig. \ref{fig: exp-results-heatmap} show that the network accurately predicts the average scaling value for horizons of 14 and 19 seconds, consistently with simulation results.

\begin{table}[t]
    \caption{\texttt{MSE} for the N-step prediction experiments (Sec. \ref{sec: N-step}).}
    \centering
    \begin{tabular}{lcccc}
        \toprule
                        & Class. Net. & Regr. Net. & LSTM-1 &  LSTM-2 \\
        \midrule
        \texttt{MSE} $\cdot 10^3$ & 5.44 & 8.92 & 78.6 & 71.9 \\
        \bottomrule
    \end{tabular}
    \label{tab: exp-results-lstm}
\end{table}

\section{Conclusions and Future Work}\label{sec:conclusions}

We have presented a learning-based approach to predict the robot's safety-induced slowdowns in Human-Robot Collaborative systems. 
We have demonstrated that a simple feed-forward neural network is effective in estimating the robot’s safety scaling factor in a real-world scenario.
As future work, we will integrate this safety scaling prediction into decision-making algorithms to dynamically select the most suitable robot task based on real-time safety and efficiency requirements.

\bibliographystyle{IEEEtran}
\bibliography{bib,bib_new}

\end{document}